\pdfoutput=1

\documentclass[]{./meta_template/fairmeta}

\usepackage{latexsym}

\usepackage{enumitem}

\usepackage[T1]{fontenc}

\usepackage[utf8]{inputenc}

\usepackage{microtype}

\usepackage{inconsolata}

\usepackage{graphicx}

\usepackage{mdframed}
\usepackage{xcolor}
\definecolor{prompt_background}{HTML}{f4faed}
\usepackage{svg}
\usepackage{array}
\usepackage{stfloats}
\usepackage{booktabs}
\usepackage{tcolorbox}
\usepackage{listings}
\usepackage{caption}
\usepackage{subcaption}
\usepackage{multirow}
\usepackage{float}
\usepackage{threeparttable}
\usepackage{bold-extra}

\renewcommand{\title}[1]{\newcommand{\titlelist}{ {\huge\bfseries #1}}}
\title{\textsc{PrismRAG}: Boosting RAG Factuality with Distractor Resilience and Strategized Reasoning}

\author{
 {Mohammad Kachuee\textsuperscript{1}},
 {Teja Gollapudi\textsuperscript{1}},
 {Minseok Kim\textsuperscript{1}},
 {Yin Huang\textsuperscript{1}},
 {Kai Sun\textsuperscript{1}},
{Xiao Yang\textsuperscript{1}},
 {Jiaqi Wang\textsuperscript{1}},
 \\
 {Nirav Shah\textsuperscript{1}},
 {Yue Liu\textsuperscript{1}},
 {Aaron Colak\textsuperscript{1}},
 {Anuj Kumar\textsuperscript{1}},
 {Wen-tau Yih\textsuperscript{2}},
 {Xin Luna Dong\textsuperscript{1}}
}

\affiliation[1]{Meta Reality Labs}
\affiliation[2]{Meta FAIR}

\date{\today}
\correspondence{\email{mkachuee@meta.com}}

\abstract{
Retrieval-augmented generation (RAG) often falls short when retrieved context includes confusing semi-relevant passages, or when answering questions require deep contextual understanding and reasoning. We propose an efficient fine-tuning framework, called {\sc PrismRAG}, that (i) trains the model with distractor-aware QA pairs mixing gold evidence with subtle distractor passages, and (ii) instills reasoning-centric habits that make the LLM plan, rationalize, and synthesize without relying on extensive human engineered instructions. Evaluated across 12 open-book RAG QA benchmarks spanning diverse application domains and scenarios, {\sc PrismRAG} improves average factuality by 5.4\%, outperforming state-of-the-art solutions. 
}

\begin{document}

\maketitle

\section{Introduction}
Factual question answering (QA) is an important application for large language models (LLMs). 
However, LLMs lack the necessary knowledge to answer questions that require current or external information not present in their parametric knowledge. To address this, \textit{retrieval-augmented generation (RAG)} is commonly employed to incorporate grounding context into the prompt~\citep{lewis2020retrieval}.
The grounding context is usually organized as reference documents fetched by search, semantic retrieval, or knowledge tools~\citep{yang2024crag,friel2024ragbench,kachuee2025improving}. In such settings, the LLM is instructed to utilize information presented in the references in conjunction with its parametric knowledge to deliver the most accurate responses~\citep{huang2025confqa}.



\begin{figure}[t]
    \centering
        \includegraphics[width=0.7\linewidth]{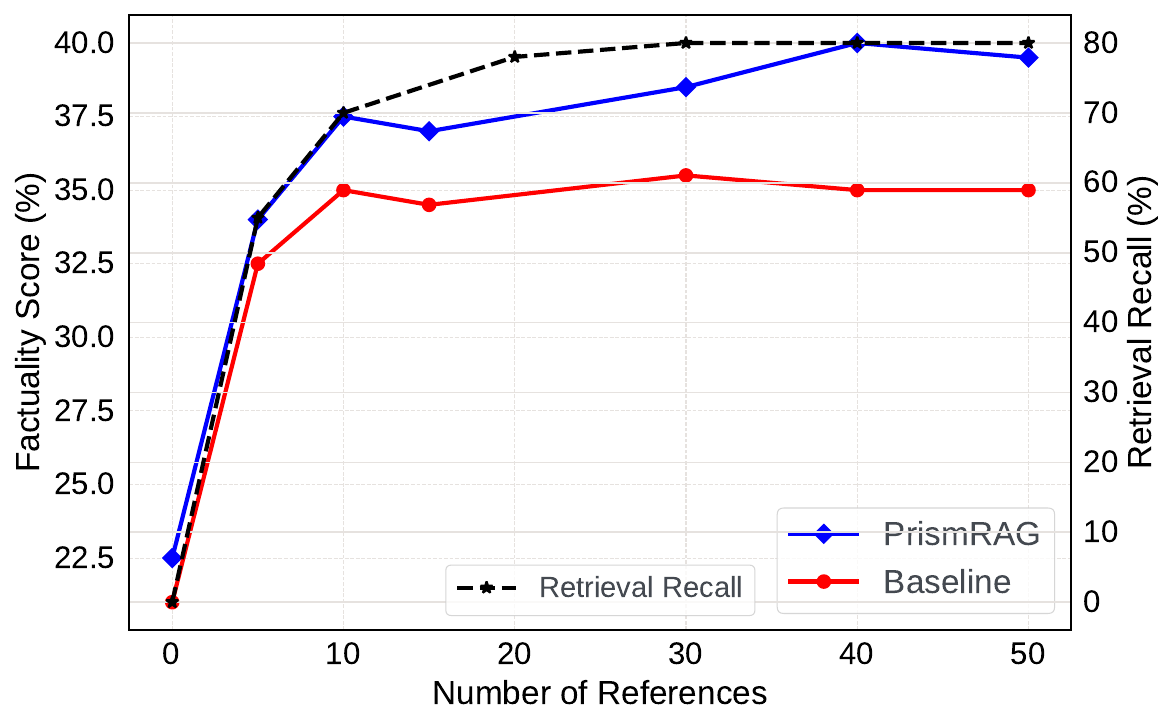}
        \caption{On the CRAG~\citep{yang2024crag} benchmark, as we increase retrieval pages from 0 to 50, whereas the retrieval recall increases to 80\%, answer factuality score (computed by accuracy minus hallucination rate) flattens out after 10 pages for \texttt{Llama-3.1-70b-instruct}. {\sc PrismRAG} improves over baseline by 4.5\%, allowing for further QA improvement as we retrieve more pages.}
        \label{fig:impact_of_ref_count_factuality}
\end{figure}

Despite its intuitive appeal, RAG still struggles to deliver reliable answers. Prior studies show that appending large blocks of retrieved content, such as entire web pages or lengthy documents, can overwhelm the model and induce hallucinations~\citep{fang2024enhancing}. 
The deficit becomes acute for nuanced questions that demand synthesizing evidence scattered across several sources.  
As revealed in Figure~\ref{fig:impact_of_ref_count_factuality}, expanding the number of context webpages from 5 to 50 boosts retrieval recall by $\sim 25\%$, yet the factuality curve plateaus, underscoring the limited benefit of ever-larger context windows.

In this paper, we propose {\sc PrismRAG}, an approach that enhances LLM's capability in generating answers from multiple retrieval results. To mitigate the negative impact of semi-relevant content retrieved alongside useful passages, we develop methods for generating training data targeted at building robustness against such distractors. Furthermore, to address the reasoning challenges inherent in RAG QA---such as assessing the relevance of retrieved passages, resolving inconsistencies, and aggregating information---we fine-tune the LLM to strategize and deliberate before producing answers. Unlike approaches that rely heavily on extensive chain-of-thought (CoT) prompt engineering, our method directly improves the model’s reasoning competence and generalization.
To summarize, we make three contributions.
\begin{enumerate}
    \item We propose {\sc PrismRAG}, a fine-tuning approach that teaches LLMs to 
    be robust against retrieval noises, and to plan and rationalize for improved answer generation.
    \item We propose a training data generation framework that combines synthetic data generation with LLM-based verification, to generate high-quality training data in a scalable fashion.
    \item We conducted comprehensive empirical study on 12 benchmarks, showing that {\sc PrismRAG} improves over baseline by 5.4\%, and outperforms state-of-the-art solutions. 
\end{enumerate}

\section{Related Work}
Since the early breakthroughs in LLMs, RAG QA has emerged as a significant application area. This approach offers a streamlined alternative to traditional web search and research for finding relevant answers~\citep{lewis2020retrieval}. Additionally, RAG QA systems facilitate the development of domain-specific applications, supporting use cases such as product user manuals, legal documents, or financial reports to name a few~\citep{sadat2023delucionqa,hendrycks2021cuad,chen2021finqa}.

RAG QA is an active area of research with multiple dimensions, including information retrieval~\citep{friel2024ragbench,kachuee2025improving}, LLM prompting~\citep{asai2023self,semnani2023wikichat},  LLM fine-tuning~\citep{cai2024forag,gekhman2024does,lin2024flame}, evaluation~\citep{yang2024crag}, and guard-railing~\citep{
kim2024groundial}.

More relevant to this work is the research on enhancing the core capabilities of LLMs through fine-tuning, aimed at addressing task complexities.
For instance, \citet{fang2024enhancing} evaluated the impact of retrieval noise on the answer quality. They categorized the noise into three types: irrelevant, relevant, and counterfactual. Their findings indicate that relevant and counterfactual noises are most detrimental. RobustRAG~\citep{xiang2024certifiably} is another example which proposed building robustness by an isolate-then-aggregate strategy.

From another perspective, recent literature highlights CoT reasoning as a powerful capability for enhancing the response quality. Reasoning helps decompose complex problems into smaller but more manageable parts, facilitating advanced contextual understanding. The current CoT literature primarily focuses on reasoning for math, logic, and decision-making, while reasoning in RAG QA remains a less explored area~\citep{wangchain,zheng2023take,phan2023training,zhang2024chain}. 
\citet{wang2025chain} proposed leveraging multi-step reasoning to dynamically query, retrieve, and evaluate documents, thereby enhancing retrieval processes.
\citet{zhang2024raft} introduced retrieval augmented fine-tuning (RAFT) and showed the value of fine-tuning on the RAG QA task, especially for CoT style responses.
Another example is LLMQuoter~\citep{bezerra2025llmquoter}, a recent method that involves fine-tuning to produce quotes before generating the answer. 

\section{Problem Definition}
We consider the problem of QA given a set of retrieved documents. Specifically, we focus on enhancing core capabilities of LLMs through fine-tuning to best leverage the internal knowledge, common sense, and reasoning capabilities in conjunction with a set of retrieved documents to produce the most factual answers. Formally, given a question $Q$, LLM parameters $\theta$, a set of retrieved documents $D=\{d_1 \dots d_n \}$, and any relevant contextual information (e.g., user time and location) $C$, the generative model generates an answer $A$:
\begin{equation}
    A = \mathcal{G}(Q, D, C | \theta) ,
\end{equation}
where $\mathcal{G}$ is used to represent the generative process. The objectives are: $(i)$ to produce an answer that addresses the question while being grounded on information present in $D$ and $C$; $(ii)$ to refrain from answering the question if the available knowledge and context do not provide sufficient information.

\section{Proposed Method}

In this section, we present our fine-tuning approach to enhance LLM's answer generation capability in presence of retrieval noises. Our approach is based on two key intuitions. First, although LLMs in general can decide the relevancy between the question and retrieved content, they may make mistakes for nuances such as for events with different dates and locations. We can enhance this capability with tailored fine-tuning (Section~\ref{sec:distractors}). Second, RAG answer generation is reasoning heavy, deciding which retrieved passages are relevant, identifying inconsistencies among retrieved content, and aggregating information across passages when necessary. Thus, we shall fine tune LLM's reasoning capability for generating better answers (Section~\ref{sec:reasoning}). These two methods improve RAG summarization from two complementary and orthogonal dimensions.

As we generate fine-tuning data, we face a dilemma. On the one hand, relying on human annotators to produce high-quality data is expensive and not scalable; this is especially true since RAG retrieval results  typically contain many pages of unstructured (e.g., long concatenated text chunks) or semi-structured (e.g., parsed tables) text, as well as structured data blobs (e.g., json output from knowledge graphs), not suited for human readers~\citep{sun2025does}. On the other hand, synthetic training data may not exhibit high quality, and noisy labels may even encourage more hallucinations. We describe in this section a simple yet effective solution for automatically producing high-quality seed data (Section~\ref{sec:seed}), and how we use LLMs to iteratively generate examples and check example quality (Section~\ref{sec:distractors}-\ref{sec:reasoning}).


\subsection{Seed QA Data}
\label{sec:seed}
We start with generating seed question-answer-passage triplets. To scale up the process, we
generate synthetic QA pairs from a given set of raw documents. Initially, we split documents into references, and then randomly select a ``golden'' reference to design a QA pair. Solving this reverse problem is a more straightforward task and can be accomplished using a reasonably strong LLM. This approach helps to produce high-quality data at scale, as the primary limiting factors would be access to raw content and compute.

We generate QA pairs using two main content sources. $(a)$ Wiki Pages, content sampled from the English Wikipedia dataset\footnote{https://dumps.wikimedia.org}; $(b)$ Web Search, parsed web pages retrieved using web search queries in response to a set of internal factual questions (details in Appendix~\ref{sec:appendix_seed_data_gen}).

\subsection{Improving Resilience to Distractors}
\label{sec:distractors}

Recent studies have highlighted the detrimental impact of distractor content~\citep{fang2024enhancing}.
To address this challenge, we propose fine-tuning on distractor-augmented content. By incorporating semi-relevant or confusing information into the training data, our approach aims to enhance the model's resilience to relevant retrieval noise thereby mitigating the risk of hallucinations and improving overall factuality. 

\begin{figure}[t]
    \centering
        \includegraphics[width=0.9\linewidth]{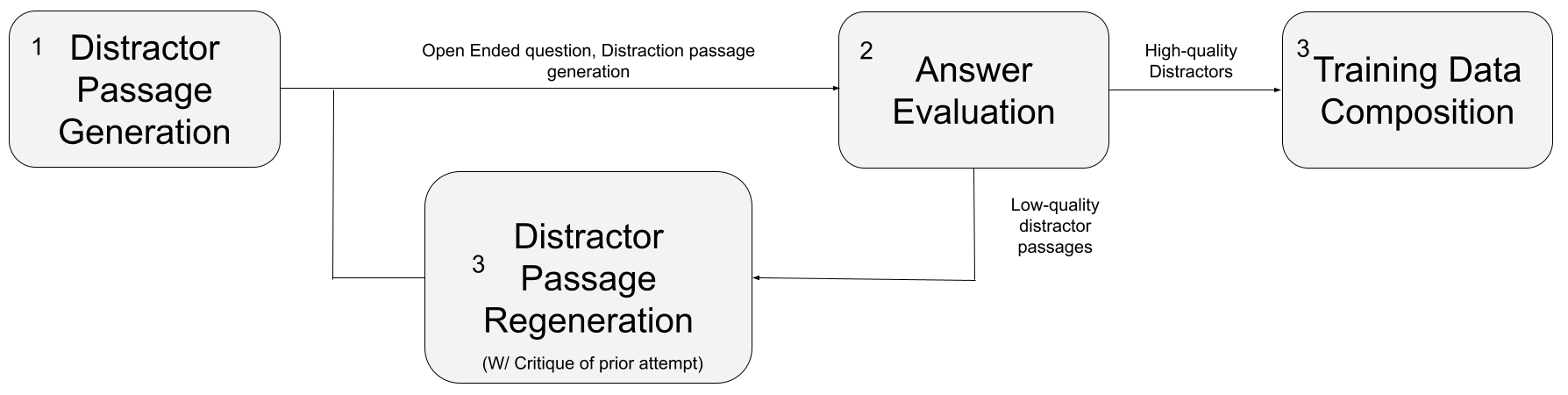}
        \caption{Overview of the synthetic distractor generation process.}
        \label{fig:distractor_flow}
\end{figure}

Informed by observations from real-world RAG QA applications and based on the seed data from Section~\ref{sec:seed}, we adopt a synthetic approach to generate targeted distractor content. This method allows for large-scale generation and precise control over the introduction of relevant noise on named-entities and temporal information.
We hypothesize that while direct solutions, such as utilizing irrelevant passages from a text corpus \citep{zhang2024raft} as distractor content add retrieval noise, they do not ensure that the noise is relevant. Through targeted synthetic generation of distractions via modifying the golden passage, we introduce retrieval noises that are specifically tailored to challenge named entities and temporal information, which our observations have shown to be particularly problematic for summarization models. 
As depicted in Figure~\ref{fig:distractor_flow}, this process consists of three steps.
\begin{enumerate}
    \item \textbf{Synthetic distractor generation:}
    Given a question, user context 
    ground-truth answer, and its grounding passage, we synthetically generate a distractor passage and an open-ended question
    (prompt in Appendix~\ref{sec:appendix_prompts_distractor_generation}).
    Specifically, we conduct three steps:
    \begin{enumerate}
        \item Identifying key entities, locations, and temporal expressions in the golden passage that are most relevant.

        \item Reformulating the original question into a more open-ended form, such that a plausible answer can be extracted from both the golden and distractor passages. 

        \item Systematically altering the identified named-entities, locations, or temporal information to create a distractor passage that is grammatically coherent, while being stylistically similar to the original.
    \end{enumerate}

    \item \textbf{Critique the effectiveness of synthetic distractor passage:}
We employ 
an evaluation prompt to assess the effectiveness of the generated distractor passages on a scale of $1$ to $5$ 
(details in Appendix~\ref{sec:appendix_prompts_dist_critq}). This step is crucial in ensuring that the distractors are relevant, confusing, and well-formatted.

\item \textbf{Train Data Composition:}
For each sample, we iterate on the above mentioned steps until each sample scores 4 or higher or reaches five iterations. For samples that pass the quality bar, we compose the training instances by combing golden passage and the generated distractor passage as references for the task.

\end{enumerate}

\subsection{Improving Reasoning and Generalization}
\label{sec:reasoning}
Numerous studies have highlighted the advantages of CoT fine-tuning, as opposed to directly training on the final answers~\citep{kim2023cot}. We believe this is particularly crucial for the QA task, where directly training on the final answer may increase risk of hallucinations due to a knowledge mismatch 
with the pre-training
~\citep{gekhman2024does}.

%
Nonetheless, there are two main considerations related to CoT fine-tuning. First, using human annotators to prepare reasoning data is challenging, as authoring rationales is not a well-defined task. Second, as reported in the literature~\citep{cheng2024chainlm,liu2024mind} and confirmed by our experiments (Appendix~\ref{sec:appendix_cot_instr_sensitivity}), CoT instructions significantly impact the quality of the final answer. 
However, prompt engineering is resource-intensive, and optimizing instructions for each domain or benchmark is neither scalable nor optimal.

\begin{figure}[t]
    \centering
        \includegraphics[width=0.9\linewidth]{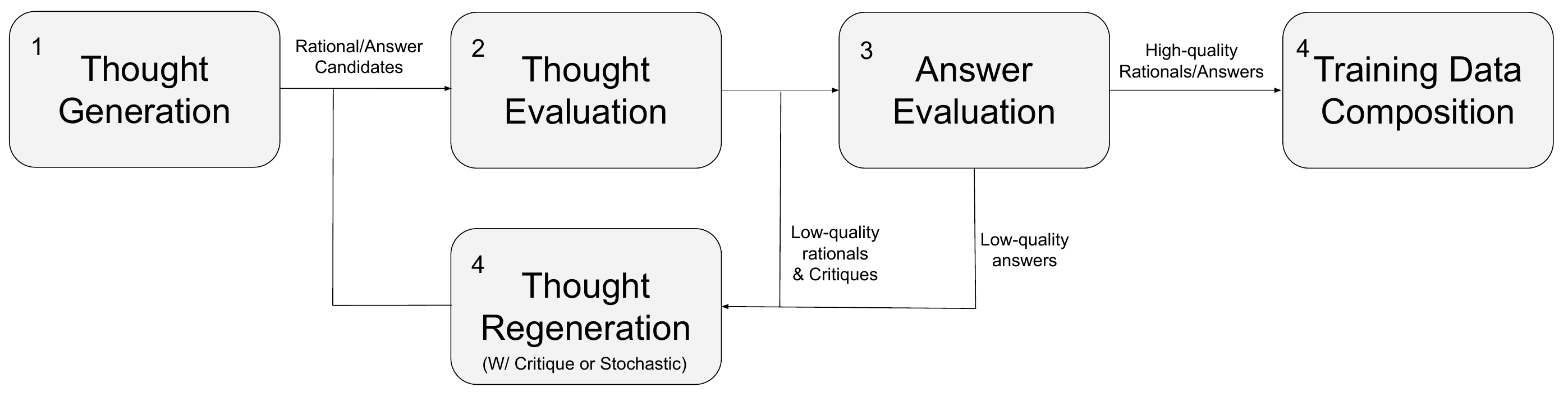}
        \caption{Overview of the iterative synthetic CoT generation process.}
        \label{fig:rethink_flow}
\end{figure}

To address these challenges, we propose strategization as a meta-task aimed at reducing reliance on CoT instruction engineering while enhancing  the performance. In this approach, the generative model is first tasked with producing a reasoning strategy that outlines the necessary thought steps. It then follows this strategy to generate the CoT and the final answer.
We hypothesize this approach teaches the model \textit{``how to think''} rather than \textit{``what to think''}, which in turn improves its ability to tackle different problem settings by dynamically adjusting CoT steps. See Figure~\ref{fig:strageization_example_a} for an example (additional examples in Appendix~\ref{sec:qualitative_strategization_examples}). As illustrated in Figure~\ref{fig:rethink_flow}, this
process consists of:
\begin{figure}[t]
    \centering
        \includegraphics[width=0.82\linewidth]{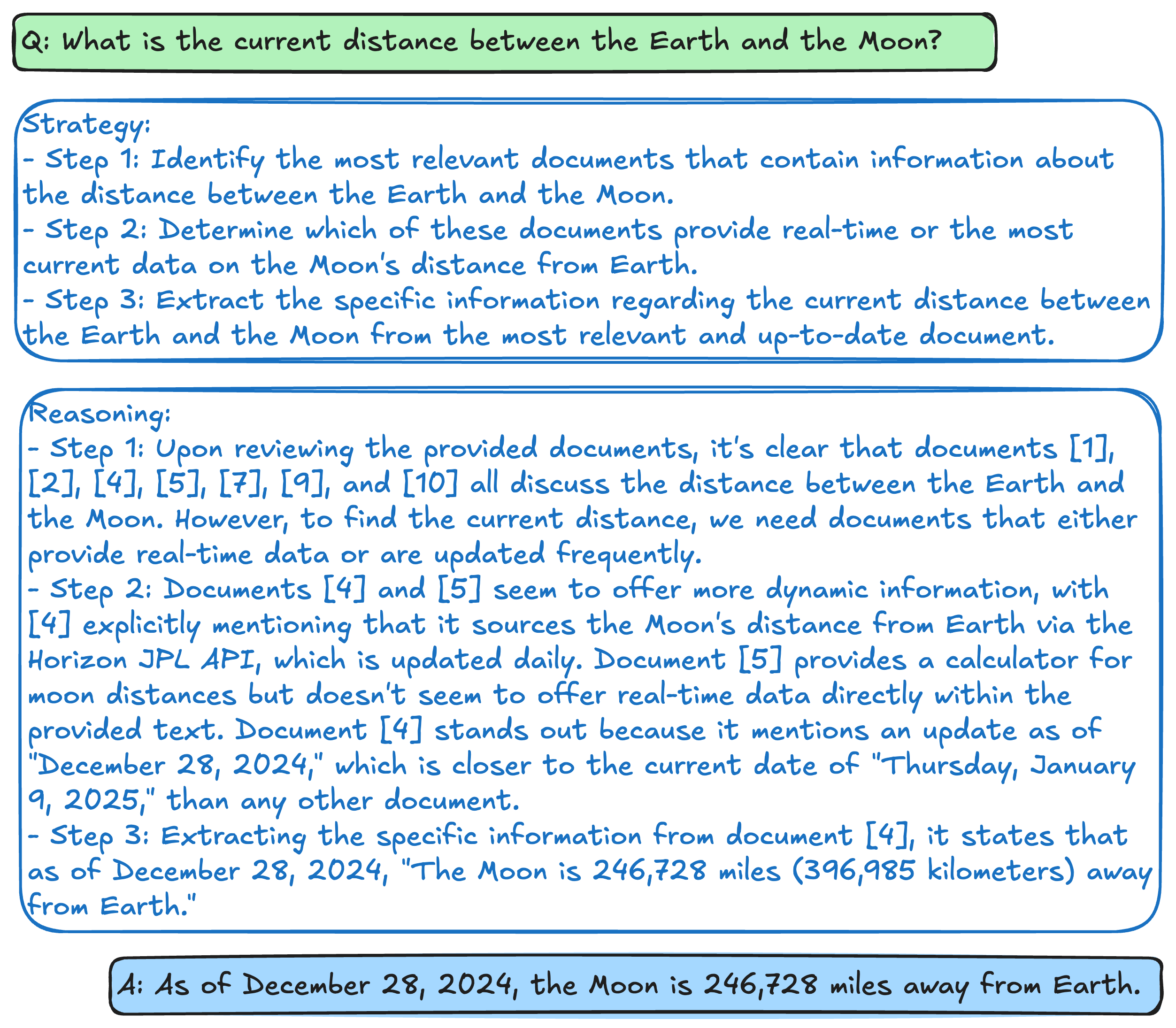}
        \caption{An example of strategization CoT inference process demonstrating the dynamic generation of CoT steps. Reference documents are not shown here.}
        \label{fig:strageization_example_a}
\end{figure}

\begin{enumerate}
    \item \textbf{Thought Generation:}
In the initial thought generation pass, we use an instruction prompt that takes the question and a set of references, and produces a strategy that outlines the necessary steps. This strategy guides the reasoning process and leads to the final answer (see the prompt in Appendix~\ref{sec:appendix_prompts_strategization}).
    \item \textbf{Thought Evaluation:}
Subsequently, we reason about the quality of the generated rationale, its thought process, and how it leads to the final answer. Then, we assign a score of $1$ to $4$, $4$ indicating excellent reasoning that leads to the ground-truth answer (prompt in Appendix~\ref{sec:appendix_prompts_thought_eval}).
We consider any samples with score below 4 to require further iterations. 
    \item \textbf{Answer Evaluation:}
We are exclusively interested in rationales that lead to the correct answer. For thoughts assessed as high-quality, we compare the candidate and the ground-truth answers to ensure the question is being fully answered and verify the factual consistency. We use a scoring system of $1$ to $4$ as the previous step
(prompt in Appendix~\ref{sec:appendix_prompts_answer_eval}). 
    \item \textbf{Thought Regeneration:}
To regenerate thoughts, we consider a combination of stochastic generation and critique-based revision methods. For stochastic generation, we simply rerun the inference in step 1 (temperature=$1.0$ and top-p=$0.90$). However, for revision, we use analysis produced during the thought evaluation step as critique feedback (prompt in Appendix~\ref{sec:appendix_prompts_rag_rethink_cot}).
    \item \textbf{Train Data Composition:}
We follow the iterative steps above until we reach a high-quality synthetic rationale, or exhaust a budget of $10$ attempts. We use stochastic thought regeneration for the first $6$ attempts, and then switch to critique-based revision for the rest. 
\end{enumerate}

\paragraph{Final training data:}
Table~\ref{tab:datamix} presents the breakdown of the final training data, consisting of a mix of distractor resilience and dynamic strategization.

\begin{table}[t]
\centering
\small
\begin{tabular}{lccc}
\toprule
\textbf{Method} & \textbf{Seed Data} & \textbf{Samples} & \textbf{Avg. Refs.} \\
\midrule
Distractor Resilience & Web Search & 2,589 & 2 \\
Dynamic Strategization & Web Search & 2,674 & 9.1 \\
Dynamic Strategization & Wiki Pages & 5,079 & 4.6 \\
\midrule
Total & & 10,342 \\
\bottomrule
\end{tabular}
\caption{Breakdown of the final training data mix.}
\label{tab:datamix}
\end{table}

\section{Experiments}

\begin{table}[t]
\centering
\resizebox{\linewidth}{!}{
\small
\begin{tabular}{l|cc|ccc|c}
\toprule
\textbf{Benchmark} & \textbf{Baseline} & \textbf{NaiveSFT} & \textbf{STaR} & \textbf{LLMQuoter} & \textbf{RAFT} & \textbf{{PrismRAG}} \\
                   &                   &                   & {\tiny\citep{zelikman2022star}} & {\tiny (Bezerra et al., 2025)} & {\tiny\citep{zhang2024raft}} & (This Work) \\
\midrule
CRAG{\tiny\citep{yang2024crag}} & 34.2\% & 27.8\% & 37.2\% & 34.4\% & 34.3\% & \textbf{39.2\%} \\
CovidQA{\tiny\citep{moller2020covid}} & 80.0\% & 83.0\% & 76.0\% & 89.0\% & 90.0\% & \textbf{95.0\%} \\
DelucionQA{\tiny\citep{sadat2023delucionqa}} & 89.0\% & 90.0\% & 92.0\% & 89.0\% & 92.0 & \textbf{97.0\%} \\
Emanual{\tiny\citep{nandy2021question}} & 92.0\% & 91.0\% & 91.0\% & 91.0\% & 92.0\% & \textbf{98.0\%} \\
ExpertQA{\tiny\citep{malaviya2023expertqa}} & 83.0\% & 83.0\% & \textbf{84.0\%} & 82.0\% & 83.0\% & 83.0\% \\
FinQA{\tiny\citep{chen2021finqa}} & \textbf{83.0\%} & 68.0\% & 72.0\% & \textbf{83.0\%} & 75.0\% & 71.0\% \\
HAGRID{\tiny\citep{kamalloo2023hagrid}} & 89.0\% & 89.0\% & 83.0\% & 89.0\% & 87.0\% & \textbf{90.0\%} \\
HotpotQA{\tiny\citep{yang2018hotpotqa}} & \textbf{93.0\%} & 63.0\% & 58.0\% & 92.0\% & 90.0\% & 89.0\% \\
MS Macro{\tiny\citep{nguyen2016ms}} & \textbf{82.0\%} & 76.0\% & 81.0\% & 81.0\% & 81.0\% & \textbf{82.0\%} \\
PubMedQA{\tiny\citep{jin2019pubmedqa}} & 80.0\% & 78.0\% & 76.0\% & 77.0\% & 78.0\% & \textbf{90.0\%} \\
TAT-QA{\tiny\citep{zhu2021tat}} & 77.0\% & 66.0\% & 66.0\% & 79.0\% & 90.0\% & \textbf{90.0\%} \\
TechQA{\tiny\citep{castelli2019techqa}} & 58.0\% & 62.0\% & 58.0\% & 75.0\% & 79.0\% & \textbf{82.0\%} \\
\midrule
Avg. & 78.4\% & 73.1\% & 72.9\% & 80.1\% & 80.9\% & \textbf{83.8\%} \\
\bottomrule
\end{tabular}
}
\caption{Comparison of RAG QA factuality scores for different approaches. {\sc PrismRAG} improves average factuality by 5.4\% over the baseline, and it outperforms SOTA such as STaR, LLMQuoter, and RAFT by $\sim 3\%-11\%$.}
\label{tab:factuality_evals}
\end{table}

\subsection{Experiment Settings}
\subsubsection{Benchmarks}
In our experiments, we employed a diverse set of 12 public RAG QA benchmarks, covering various question domains, answer formats, and reference document types. Specifically, we used the web portion of the CRAG dataset~\citep{yang2024crag}, after filtering out the false-premise type of questions, we utilized BGE embedding~\citep{xiao2024c} to rank references, resulting in a total of 643 samples.

For the remaining 11, we leveraged pre-processed data from RAGBench~\citep{friel2024ragbench}, down-sampling each to 100 samples. These benchmarks cover relevant real-world application areas, including health~\citep{jin2019pubmedqa,moller2020covid}, finance~\citep{chen2021finqa,zhu2021tat}, customer support~\citep{sadat2023delucionqa,nandy2021question,malaviya2023expertqa}, legal~\citep{hendrycks2021cuad}, and general knowledge~\citep{yang2018hotpotqa,kamalloo2023hagrid,malaviya2023expertqa}. 

\subsubsection{Metrics}
To evaluate the factual quality of generated answers, we follow 
metrics suggested by~\citet{yang2024crag}, classifying each response as either accurate, hallucinated, or missing. Accurate responses fully and accurately answer the question. Hallucinated answers contain inaccurate or misleading information. Missing responses either refuse to answer or fail to completely address the question. To summarize the overall factuality, we use a \textit{factuality score}, defined as the accuracy rate minus the hallucination rate. Factuality score provides a scalar measure of overall answer quality, where hallucinated answers are penalized twice as much as missing answers.

To evaluate factuality for CRAG, HotpotQA, MS Marco, FinQA, TAT-QA, HAGRID, and ExpertQA 
, we employed an LLM-as-judge to compare the ground-truth with the candidate answers. For other benchmarks, we utilized a fact-checking tool similar to VeriScore~\citep{song2024veriscore}, which is more suitable for long-form answers or scenarios with multiple potential answers.

\subsubsection{Implementation}
We used \texttt{Llama-3.1-70b-instruct}~\citep{grattafiori2024llama} as the base model across all our experiments. For fine-tuning, we experimented with learning rate of $10^{-5}$, computing loss over the entire assistant response (strategy, thought, and answer when present) but not using loss over the instruction prompt. For inference, we use a typical RAG prompt as in Appendix~\ref{sec:appendix_prompts_rag_simple_cot} with generation temperature of $1.0$ and top-p of $0.9$.

For comparisons with other work, to ensure fairness, we leveraged the same seed data and dataset size, while re-implementing their train data generation logic. Specifically, for: $(a)$ \textbf{NaiveSFT}: Directly trained on final answers without strategization and CoT. $(b)$ \textbf{STaR}~\citep{zelikman2022star}: Implemented a rationalization chain for RAG QA using static CoT instructions. $(c)$ \textbf{LLMQuoter}~\citep{bezerra2025llmquoter}: Implemented the same quote extraction instructions to produce training data. $(d)$ \textbf{RAFT}~\citep{zhang2024raft}: Reused their data generation code-base replacing closed-sourced model endpoints with LLaMA.


\subsection{Results}


\paragraph{Overall results:}
Table~\ref{tab:factuality_evals} presents a comparison of the proposed method ({\sc PrismRAG}) with the baseline model and other related work in the literature.
As shown in this Table, the proposed method demonstrates substantial improvements in factuality across benchmarks. Notably, {\sc PrismRAG} achieves a factuality score improvement of $5\%$ for the CRAG dataset, delivering best results in $9$ out of $12$ benchmarks, and an overall macro-average gain of $5.4\%$ over the baseline. From the average results, NaiveSFT and STaR regress over the baseline, while LLMQuoter and RAFT show more promise. This finding confirms the drawbacks of directly training on QA labels, while highlighting the benefits of training on intermediate tasks (e.g., quote extraction in LLMQuoter) and training for resilience (e.g., irrelevant references in RAFT).
The breakdown of results is provided in Appendix~\ref{sec:appendix_factuality_breakdowns}.

\paragraph{Sensitivity to references:} 
We conducted an experiment using the CRAG benchmark by limiting the number of references and measuring the impact on performance. From Figure~\ref{fig:impact_of_ref_count_factuality} results, the proposed method consistently outperforms the baseline, with its margin of improvement increasing as more references are used. This demonstrates the effectiveness of {\sc PrismRAG} in utilizing retrieved documents (see breakdown charts in Appendix~\ref{sec:appendix_impact_of_ref_count_factuality}).

\paragraph{Ablation study:} Table~\ref{tab:crag_results} presents an ablation study examining the impact of two major components of the proposed fine-tuning method: distractor resilience and dynamic strategization. From this analysis, strategization contributes to increased accuracy and reduced hallucinations, while the distractor resilience task specifically aids in reducing hallucinations. The combination of both methods yields the best overall factuality results, highlighting their complementary roles and effectiveness.

Additional experiments for the closed-book settings are provided in Appendix~\ref{sec:appendix_closed_book_qa}.

\begin{table}[t]
\centering
\small
\begin{tabular}{l|cccc}
\toprule
\textbf{Method} & \textbf{Accurate} & \textbf{Hallucinated} & \textbf{Missing} & \textbf{Factuality} \\
\midrule
Baseline & 59.1\% & 24.9\% & 16.0\% & 34.2\% \\
\hline
{\sc PrismRAG} & 62.1\% & 22.9\% & 15.1\% & 39.2\% \\
{\small -Distractor} & 59.3\% & 23.2\% & 17.6\% & 37.0\% \\
{\small -Strategization} & 62.4\% & 23.2\% & 12.3\% & 36.1\% \\
\bottomrule
\end{tabular}
\caption{Ablation study using the CRAG dataset for the proposed fine-tuning method based on distractor resilience and dynamic strategization.}
\label{tab:crag_results}
\end{table}

\section{Conclusion}
In this paper, we tackle key challenges in open-book question answering by bolstering the robustness of LLMs against distractors and enhancing their reasoning capabilities. Our approach involved fine-tuning LLMs with synthetic distractor data and introducing a novel strategization method focused on dynamic CoT step generation. Extensive evaluations across 12 public benchmarks demonstrated significant improvements in factual accuracy. These findings highlight the potential of our approach to advance the state-of-the-art, 
ultimately delivering a more factual and reliable QA system.

\section*{Limitations}
While our proposed method shows significant improvements in factuality for open-book question answering, several limitations persist. First, the reliance on synthetically generated distractor data may not fully capture the complexity and variability of real-world distractors, potentially limiting the model's robustness in diverse scenarios. Additionally, the use of LLM-as-judge for factuality evaluation, both in our training data pipeline and benchmarking, presents its own challenges. For instance, it may exhibit bias towards the presence or absence of additional explanations, and its behavior can become less predictable when there is a mismatch between its internal parametric knowledge and the retrieved references.

\clearpage
\newpage
\bibliographystyle{meta_template/assets/plainnat}
\bibliography{refs}
\clearpage
\newpage
\beginappendix

\section{Sensitivity to CoT Instructions}
\label{sec:appendix_cot_instr_sensitivity}
To study the sensitivity of RAG QA generated answers to prompt instructions, we experimented with three variations of the instruction prompt in~\ref{sec:appendix_prompts_rag_simple_cot} (Prompt-A). For the first variant (Prompt-B), updated step 1 to directly ask for evaluating individual references, and consolidated step 2 and 3 with emphasis on saying ``I don't know'' when provided information is insufficient. For the second variant (Prompt-C), we removed the emphasis on ``I don't know''. Finally, in the last variant (Prompt-D), we removed all CoT step instructions and just asked for thinking before answering the question. In these experiments, we used \texttt{llama-3.3-70b-instruct}~\citep{grattafiori2024llama} as the generative model and a holdout set from the Web Pages (\ref{sec:appendix_seed_data_gen}) dataset as our test set.

Based on the comparison of results presented in Table~\ref{tab:cot_instr_sensitivity}, alight the base instructions are the same, and the only difference is CoT step instructions, we see significant sensitivity to the prompt variations. This is particularly important for the case of refusal answers as explicit step instruction on refusing with ``I don't know'', seem to severely increase the refusal rate, while a very similar instruction is provided in the task definition above CoT steps for the base prompt. Also, it is worth noting that having no instructions, and simply asking to "think before answer" has a competitive performance compared to engineered prompts for this specific task.

\begin{table}[h]
\centering
\small
\begin{tabular}{l|cccc}
\toprule
\textbf{Instruction} & \textbf{Accuracy} & \textbf{Hallucination} & \textbf{Missing} & \textbf{Factuality} \\
\midrule
Prompt-A & 69.9\% &  18.4\% & 11.7\% & 51.5\% \\
Prompt-B & 63.2\% &  15.5\% & 21.4\% & 47.7\% \\
Prompt-C & 69.2\% &  18.7\% & 12.1\% & 50.5\% \\
Prompt-D & 69.3\% &  19.6\% & 11.1\% & 49.7\% \\
\bottomrule
\end{tabular}
\caption{Experiments to study the sensitivity of RAG QA factuality metrics to CoT step instructions.}
\label{tab:cot_instr_sensitivity}
\end{table}

\section{Strategization Examples}
We present two additional examples for the strategization CoT process. In the first example, the reasoning steps are dynamically defined to focus on specific sections and entities in a legal document to extract the final summary. However, in the second example, the focus is on comparing and analyzing entities relented such as language and publication year for a book title before proving the final answer.
\label{sec:qualitative_strategization_examples}
\begin{figure}[h]
    \centering
        \includegraphics[width=0.49\linewidth]{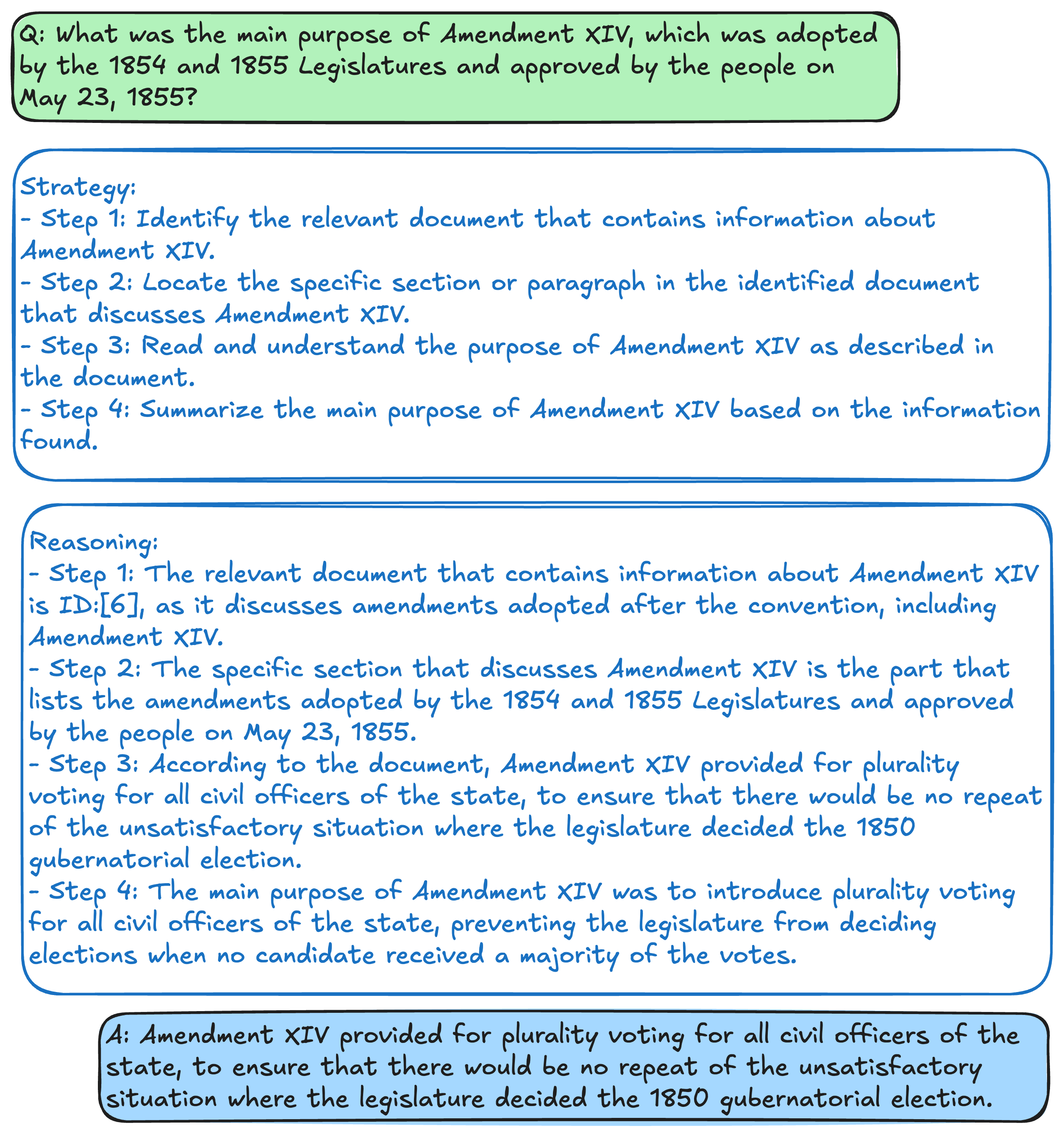}~\includegraphics[width=0.49\linewidth]{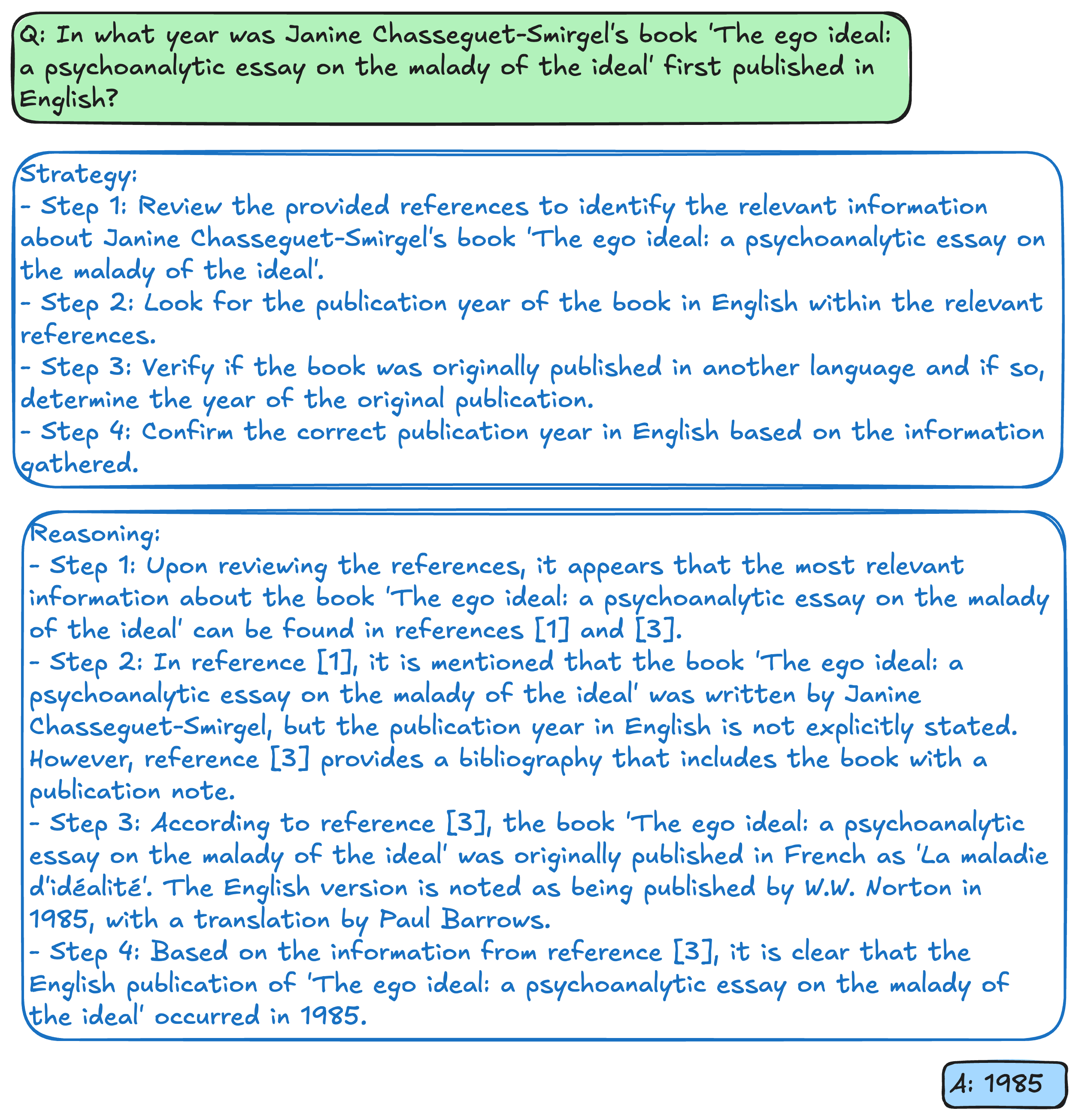}
        \caption{Examples of strategization CoT inference process demonstrating the dynamic generation of CoT steps. Samples taken from the Wiki dataset. Reference documents are not shown here.}
        \label{fig:strageization_example_b}
\end{figure}

\section{Seed Data Generation}
\label{sec:appendix_seed_data_gen}
Here, we consider two main content sources:
\begin{itemize}
    \item \textbf{Wiki Pages}: We sampled about 10K pages from the English Wikipedia data set\footnote{https://dumps.wikimedia.org}. First, filtered very short pages of less than $500$ words or $10$ lines as well as any pages longer than $7000$ words or more than $1000$ lines. Then, we split each page to non-overlapping text chunks, randomized and shuffled to eventually have 2 to 15 references of size between 250 to 1000 words. This process resulted in final sample size of about 5K RAG QA task samples.
    \item \textbf{Web Search}: To further diversify and target more challenging scenarios, we curated a collection of pages retrieved using web search queries in response to a set of internal knowledge related questions that are primarily focused on time-sensitive queries (e.g. news events, sports outcomes). For each query we retrieved up to 10 pages. Consequently, each page was converted from html to a plain text reference document that is truncated to 3000 words. For this data source, we also use and include relevant time and location (city name) of the request as additional contextual information.
\end{itemize}

To generate synthetic QA pairs for references extracted from the Wiki Pages, we randomly select a reference and leverage an instruction prompt focused on producing QA pairs that are grounded on the reference content. We found that emphasizing on question difficulty (e.g. college graduate level and setting a difficulty grade), listing a few common category of problematic questions to avoid (e.g., questions about user time/location), and providing exemplars significantly helps with the final data quality. Additionally, we instruct the model to reject designing QA when reference content is not readable, which is sometimes the case for the Web Search references due to html parse failures or web page retrieval/access problems. For the complete prompt, please refer to Appendix \ref{sec:appendix_prompts_qa_synthetic}. Regarding the LLM, we used \texttt{llama-3.3-70b-instruct}~\citep{grattafiori2024llama}\footnote{https://www.llama.com/models/llama-3/} as we found it reasonably capable and cost-efficient for this use case.

\section{Breakdown of Results}
\label{sec:appendix_factuality_breakdowns}
Table~\ref{tab:results_breakdown} presents the breakdown results for the baseline and proposed method. We follow a similar metric design as suggested by ~\citet{yang2024crag} to classify each response to one of accurate, hallucinated, or missing. Accurate responses fully and accurately answer the question. Hallucinated answers contain inaccurate and misleading facts. Missing is any answer that refuses to answer or fails to completely address the asked question. Finally, to summarize the overall factuality, we use \textit{factuality score}, defined as hallucination rate subtracted from accuracy. Note that the ``Unverified'' portion is applicable to benchmarks without provided ground-truth labels and cases where our factuality analysis tool was not able to verify correctness of all claims in the answer.

\begin{table}[h]
\centering
\begin{tabular}{llccccc}
\toprule
   & & Correct & Hallucination & Refusal & Unverified & Factuality Score \\
\midrule
\multirow{12}{*}{Baseline} & CRAG   & 59.1\% & 24.9\% & 16.0\% & 0.0\% & 34.2\% \\
& CovidQA   & 85.0\% & 5.0\% & 8.0\% & 2.0\% & 80.0\% \\
& DelucionQA & 93.0\% & 4.0\% & 3.0\% & 0.0\% & 89.0\% \\
& Emanual    & 93.0\% & 1.0\% & 2.0\% & 4.0\% & 92.0\% \\
& ExpertQA   & 86.0\% & 3.0\% & 7.0\% & 4.0\% & 83.0\% \\
& FinQA      & 89.0\% & 6.0\% & 2.0\% & 3.0\% & 83.0\% \\
& HAGRID     & 93.0\% & 4.0\% & 1.0\% & 2.0\% & 89.0\% \\
& HotpotQA   & 96.0\% & 3.0\% & 0.0\% & 1.0\% & 93.0\% \\
& MS Macro   & 90.0\% & 8.0\% & 1.0\% & 1.0\% & 82.0\% \\
& PubMedQA   & 85.0\% & 5.0\% & 7.0\% & 3.0\% & 80.0\% \\
& TAT-QA     & 87.0\% & 10.0\% & 2.0\% & 1.0\% & 77.0\% \\
& TechQA     & 61.0\% & 3.0\% & 21.0\% & 15.0\% & 58.0\% \\
\hline
\multirow{12}{*}{This Work} & CRAG   & 62.1\% & 22.9\% & 15.1\% & 0.0\% & 39.2\% \\
& CovidQA   & 96.0\% & 1.0\% & 2.0\% & 1.0\% & 95.0\% \\
& DelucionQA & 98.0\% & 1.0\% & 1.0\% & 0.0\% & 97.0\% \\
& Emanual    & 98.0\% & 0.0\% & 2.0\% & 0.0\% & 98.0\% \\
& ExpertQA   & 84.0\% & 1.0\% & 13.0\% & 2.0\% & 83.0\% \\
& FinQA      & 81.0\% & 10.0\% & 3.0\% & 6.0\% & 71.0\% \\
& HAGRID     & 91.0\% & 1.0\% & 7.0\% & 1.0\% & 90.0\% \\
& HotpotQA   & 92.0\% & 3.0\% & 0.0\% & 5.0\% & 89.0\% \\
& MS Macro   & 89.0\% & 7.0\% & 1.0\% & 3.0\% & 82.0\% \\
& PubMedQA   & 92.0\% & 2.0\% & 6.0\% & 0.0\% & 90.0\% \\
& TAT-QA     & 94.0\% & 4.0\% & 2.0\% & 0.0\% & 90.0\% \\
& TechQA     & 86.0\% & 4.0\% & 7.0\% & 3.0\% & 82.0\% \\
\bottomrule
\end{tabular}
\caption{Breakdown of factuality benchmark results.}
\label{tab:results_breakdown}
\end{table}

\section{CRAG Resilience Experiment}
\label{sec:appendix_impact_of_ref_count_factuality}

\begin{figure}[h]
    \centering
        \includegraphics[width=0.7\linewidth]{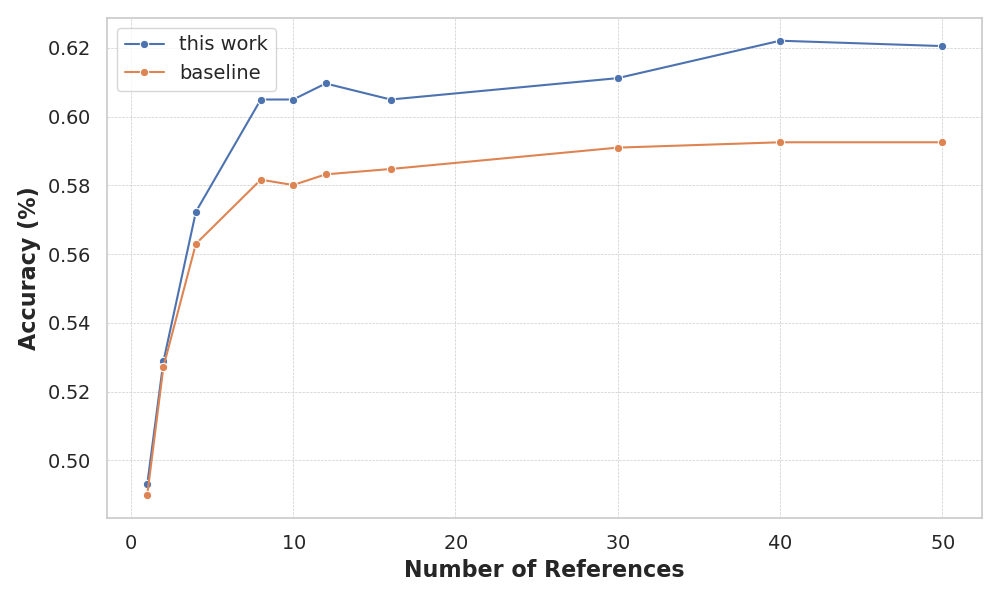}
        \includegraphics[width=0.7\linewidth]{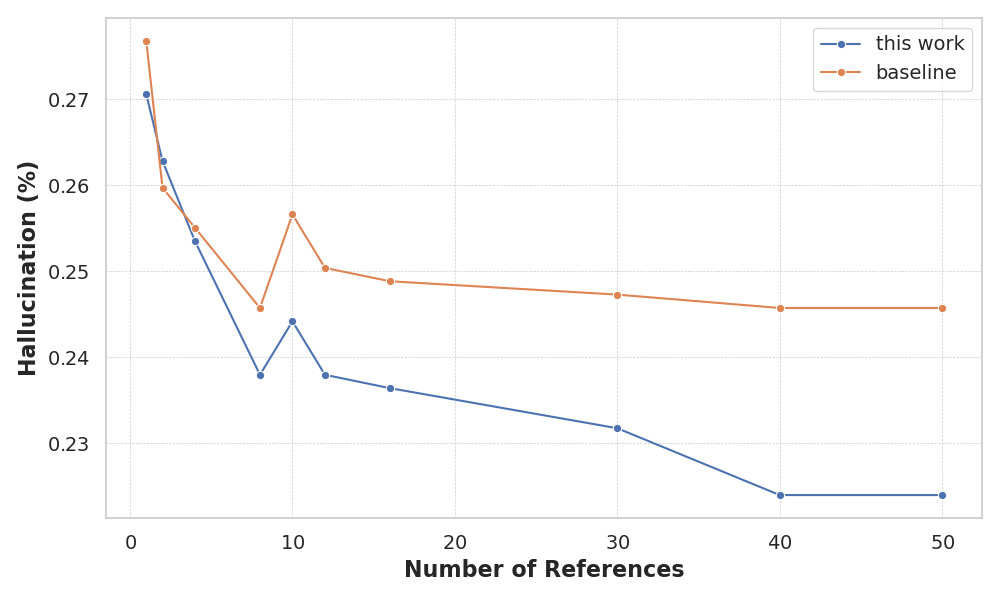}
        \includegraphics[width=0.7\linewidth]{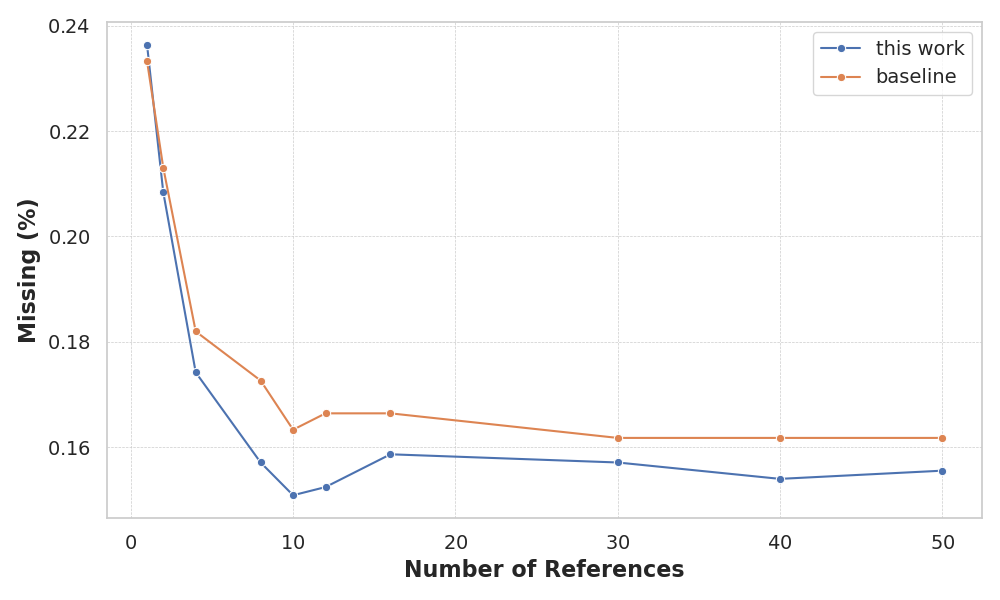}
        \caption{Impact of changing the number of reference documents on accuracy, hallucination, and missing rate for the CRAG~\citep{yang2024crag} benchmark.}
        \label{fig:breakdown_impact_of_ref_count_factuality}
\end{figure}

We conducted an experiment using the CRAG benchmark by limiting the number of reference documents and measuring the impact on the performance. From Figure~\ref{fig:breakdown_impact_of_ref_count_factuality}, the proposed method consistently outperforms the baseline. For the accuracy and hallucination rates, the margin of improvement increases with the use of more references, showing the effectiveness of the proposed method to reject noises and leverage additional grounding content.

\section{Evaluation on Closed-Book QA}
\label{sec:appendix_closed_book_qa}
\subsection{Experiment Settings}
To further demonstrate the effectiveness of our proposed approach, we conducted an additional evaluation on closed-book factuality QA benchmarks. These benchmarks contain short answers for given questions and do not have access to retrieved documents, allowing us to isolate and showcase the enhancement in reasoning capability afforded by our method.
We describe their details in the following section.
\subsubsection{Benchmarks}
\begin{itemize}
    \item \textbf{CRAG}~\citep{yang2024crag} is designed to evaluate a model's RAG capability. For the comparisons made in this section, we have used the same web summarization split from main text but without any reference documents.
    \item \textbf{SimpleQA}~\citep{wei2024measuring} comprises 4,326 short QA pairs that cover a wide range of topics, including history, technology, science, and entertainment.
    \item \textbf{Head-to-Tail}~\citep{sun2022recitation} contains simple QA pairs generated from a general knowledge graph, DBPedia, and an internet movie database, IMDb. For each dataset, we sampled 200 entities respectively from head, torso, and tail entities which are divided by entity popularity following \citep{sun2023head}, resulting in 1,200 QA pairs.
\end{itemize}
\subsubsection{Metrics and Implementation}
For short-form factuality QA, we use accuracy, missing, and hallucination following \citep{yang2024crag} followed by factuality score described in the main text. 

Identical to the evaluation in the main text, we compared \texttt{llama-3.1-70b-instruct}~\citep{grattafiori2024llama} and its fined-tuned version using our proposed approach. We note that the recipe for fine-tuning and inference is also the same with the setting used in the main text.

\subsection{Results}
As shown in Table~\ref{tab:short_form_factuality}, our proposed method demonstrates significant factuality gains across all datasets, with the exception of DBPedia for which we see a marginal regression. Notably, we observe a substantial reduction in hallucination rates across all four datasets. These findings suggest that the proposed approach can effectively enhance a model's factuality capabilities and generalization to even in closed-book settings, despite being trained on samples that utilize search documents.

\begin{table}[h]
\centering
\small
\begin{threeparttable}
\begin{tabular}{l|cccc}
\toprule
\textbf{Model} & \multicolumn{4}{c}{\textbf{CRAG}\tnote{a}~~(closed-book)} \\ \cmidrule{2-5}
               & Acc (Rec) & Miss. & Hall. & Fac. \\
\midrule
Baseline       & 58.7      & 15.6  & 25.7  & 33.0 \\
This Work      & 55.3      & 22.6  & 22.1  & \textbf{33.2} \\
\midrule
\textbf{Model} & \multicolumn{4}{c}{\textbf{SimpleQA} (closed-book)} \\ \cmidrule{2-5}
               & Acc (Rec) & Miss. & Hall. & Fac. \\
\midrule
Baseline       & 20.0      & 44.1  & 35.9  & -15.8 \\
This Work      & 14.0      & 64.0  & 22.0  & \textbf{-8.0}  \\
\midrule
\textbf{Model} & \multicolumn{4}{c}{\textbf{IMDB} (closed-book)} \\ \cmidrule{2-5}
               & Acc (Rec) & Miss. & Hall. & Fac. \\
\midrule
Baseline       & 44.8      & 34.2  & 21.0  & 23.8 \\
This Work      & 40.7      & 47.7  & 11.7  & \textbf{29.0} \\
\midrule
\textbf{Model} & \multicolumn{4}{c}{\textbf{DBpedia} (closed-book)} \\ \cmidrule{2-5}
               & Acc (Rec) & Miss. & Hall. & Fac. \\
\midrule
Baseline       & 52.0      & 22.0  & 26.0  & \textbf{26.0} \\
This Work      & 41.0      & 43.0  & 16.0  & 25.0 \\
\bottomrule
\end{tabular}
\begin{tablenotes}
\item[a] CRAG benchmark here is different from the main text as reference documents are omitted for the close-book task.
\end{tablenotes}
\end{threeparttable}
\caption{Closed-book factuality benchmark results.}
\label{tab:short_form_factuality}
\end{table}

\section{Prompts}
\label{sec:appendix_prompts}
\subsection{Synthetic QA Generation}
\label{sec:appendix_prompts_qa_synthetic}
\begin{figure*}[h!]
\begin{mdframed}[backgroundcolor=prompt_background,linewidth=0.0pt]
\texttt{\small
You are a helpful assistant. Always follow the provided instructions and generate outputs in valid json format without any extra information. Generate a question and answer pair based on the Provided Content below. \\
\\
\#\# Requirements: \\
- You must ground your question and answer to the Provided Content \\
- The question should be selected to resemble what a curious college graduate would ask an intelligent conversational system. From the difficulty level of 1 to 10, aim for an 8. \\
- The answer should be fully and directly grounded on the Provided Content. Never use any information other than what is available in the Provided Content to generate the question and answer. \\
- Never generate a question that is asking for the current time, date, or location. \\
- The question should not be too general or vague. When applicable, include specific entities, names, times, locations, events, and keywords in the question. \\
- The question and answer must be grammatically correct and be conversationally natural. \\
- A good question should be meaningful and provides enough context. For example, questions like "when was this updated?" do not provide enough context and are not meaningful, but "when was the start date for world war two?" is clear and meaningful.  \\
- Always return in json format with two keys: "question" and "answer". If the Provided Content is not readable, you may set the value corresponding to the question and answer keys to "N/A". \\
\\
\#\# Examples:\\
Here are some examples of questions types to consider: \\
1. How old is Obama? \\
2. What was the name of the first president of the United States? \\
3. How is the weather in Seattle this weekend? \\
4. What is the population of China? \\
5. Is there a movie theater nearby? \\
6. What time is high tide tonight in Santa Cruz, CA? \\
7. Who is leading in the election between Trump and Kamala Harris? \\
8. What is the dodgers current score? \\
9. When does daylight saving time start? \\
10. Any updates on Morgan Freeman's health? \\
11. What are the main ingredients in a margarita? \\
12. Why is chocolate bad for dogs? \\
\\
\#\# Provided Content:\\
<reference content>
}
\end{mdframed}
\caption{Prompt used for synthetic QA generation given reference content.}
\label{fig:qa_gen_prompt}
\end{figure*}

\subsection{Vanilla RAG QA CoT Prompt}
\label{sec:appendix_prompts_rag_simple_cot}
\begin{figure*}[h!]
\begin{mdframed}[backgroundcolor=prompt_background,linewidth=0.0pt]
\texttt{\small
For this task you are asked to answer a question (Question). Please provide factually accurate, direct, and clear responses. Do not ask any clarification questions or ask for additional information.
To make sure the right facts are being considered, you should always ground your responses to the provided references below (References). If the references are not relevant to the question or do not provide the right information, you can respond with an apology rather than fabricating facts.\\\\
\#\# References:\\
<Reference Documents>\\\\
\#\# Question:\\
<Question>\\\\
Now let's think step by step:\\
    Step 1: summarize what is the question is asking, and what are the specific key pieces of information that are needed to answer the question.\\
    Step 2: analyze the provided references, one by one. Identify the relevant information that can be used to answer the question. Pay close attention to the entities, names, times, locations, events, and keywords that are relevant to the question. If the question is related to (or implies) the current user location and/or time, you must consider that in finding the relevant information and answering the question.\\
    Step 3: based on Step 1 and Step 2, you must provide an answer that directly address the question and is fully grounded on the provided references.\\\\
    Now, provide your reasoning steps in a few sentences followed by the final answer to the question as a new line starting with "\#\# Answer:".}
\end{mdframed}
\caption{Prompt used for answer and thought generation given the question and reference content.}
\label{fig:vanilla_rag_prompt}
\end{figure*}

\subsection{Thought Evaluation Prompt}
\label{sec:appendix_prompts_thought_eval}
\begin{figure*}[h!]
\begin{mdframed}[backgroundcolor=prompt_background,linewidth=0.0pt]
\texttt{\small
For this task you are given a question (Question), a set of references (References) as well as a reasoning (Reasoning) and an answer (Answer) to the question.\\
Your task is to evaluate the reasoning process. Use the following guidelines to evaluate the reasoning process and assign a score between 1 to 4 to the provided reasoning:\\
- Score 1: The reasoning is not correct. It is not related to the question and the references, or it is a simple repetition of them with no thought process.\\
- Score 2: The reasoning is related to the question and the references, but it is not really helpful in answering the question. It is not clear how the reasoning leads to the answer.\\
- Score 3: The reasoning is related to the question and the references, and it is helpful in answering the question. It may partially help answering the question but there are gaps in the reasoning and parts of the question are not addressed.\\
- Score 4: The reasoning is related to the question and the references, and it is helpful in answering the question. It provides a clear and complete thought process that leads to the answer.\\\\
\#\# References:\\
<Reference Documents>\\\\
\#\# Question:\\
<Question>\\\\
\#\# Reasoning:\\
<Thought>\\\\
\#\# Answer:\\
<Answer>\\\\
First, explain your assessment of the Reasoning based on the criteria above in a few sentences followed by the final score in a new line starting with "\#\# Score:" followed by the score value.}
\end{mdframed}
\caption{Prompt used for evaluation of the thought quality.}
\label{fig:thought_eval_prompt}
\end{figure*}

\subsection{Answer Evaluation Prompt}
\label{sec:appendix_prompts_answer_eval}
\begin{figure*}[h!]
\begin{mdframed}[backgroundcolor=prompt_background,linewidth=0.0pt]
\texttt{\tiny For this task you are given a question (Question), a reference answer (Reference Answer), and a candidate answer (Candidate Answer).\\
Your task is to evaluate if the Candidate Answer is fully answers the question while being consistent to the information presented in the Reference Answer.\\
Use the following guidelines to evaluate the reasoning process and assign a score between 1 to 4 to the candidate answer:\\
- Score 1: The candidate answer does not provide any information to answer the question. It is a simple repetition of the question, a refusal to answer, or providing irrelevant information.\\
- Score 2: The the candidate answer is not consistent with the reference answer and there are major points of contradiction.\\
- Score 3: The the candidate answer partially answers the question and is consistent with the reference answer. It may have some minor points of contradiction but it is consistent for the most part.\\
- Score 4: The the candidate answer fully answers the question and is consistent with the reference answer.\\
Note: the reference answer may provide additional information that is not needed to answer the question. In such cases, the candidate answer does not have to contain such extra information to be considered consistent or complete, it simply has to provide a clear and complete answer to the original question while not contradicting the reference answer.\\\\
\#\# Examples:\\
Question: How tall is the Eiffel Tower?\\
Reference Answer: The Eiffel Tower in located in Paris and is about 984 feet tall.\\
Candidate Answer: It is 984 feet tall.\\
Reasoning: The candidate answer fully answers the question and is consistent with the reference answer. Extra information about the location of the Eiffel Tower is not needed to answer the question.\\
Score: 4\\\\
Question: What is the name of largest waterfall in the world and where is it located?\\
Reference Answer: The largest waterfall in the world is Angel Falls in Venezuela and is about 3,212 feet tall.\\
Candidate Answer: The largest waterfall in the world is Angel Falls.\\
Reasoning: The candidate answer parially answers the question and is consistent with the reference answer. However, it does not provide the location of the waterfall.\\
Score: 3\\\\
Question: Which country had most gold medals in the 2022 Winter Olympics?\\
Reference Answer: The country with most gold medals in the 2022 Winter Olympics was Canada\\
Candidate Answer: United States had most gold medals in that Olympics.\\
Reasoning: The candidate answer provides a different country name which is not consistent with the reference answer.\\
Score: 2\\\\
Question: What is the capital of France?\\
Reference Answer: The capital of France is Paris.\\
Candidate Answer: I don't know.\\
Reasoning: The candidate answer does not provide any information to answer the question. It is refusing to answer the question.\\
Score: 1\\
}
\end{mdframed}
\caption{Part 1: Prompt used for evaluation of the answer quality.}
\label{fig:answer_eval_prompt}
\end{figure*}

\begin{figure*}[h!]
\begin{mdframed}[backgroundcolor=prompt_background,linewidth=0.0pt]
\texttt{\tiny
Question: Who maintains traffic signals in the unincorporated areas of Collier County?\\
Reference Answer: Collier County operates and maintains 283 traffic signals in the unincorporated areas of the County, which generally includes areas north of Pine Ridge Road and east of Goodlette-Frank Road.\\
Candidate Answer: Collier County.\\
Reasoning: The candidate answer is consistent with the reference answer and provides the specific information needed to answer the question.\\
Score: 4\\\\
Question: What is included in the Footlong Quarter Pound Coney Combo at Sonic Drive-in?
Reference Answer: The Provided Content does not specify the details of the Footlong Quarter Pound Coney Combo, but it is mentioned as one of the menu items.\\
Candidate Answer: Sorry, the provided references do not contain the necessary information to answer the question.
Reasoning: The candidate answer does not provide any information to answer the question and instead saying the information is not available.\\
Score: 1\\\\
Question: What is the best-selling album of The Crystal Method in the United States?
Reference Answer: The album Vegas, which has sold more than one million copies in the United States, certifying it platinum.\\
Candidate Answer: Vegas\\
Reasoning: The candidate answer is directly answering the question and is mentioning the same album name as the reference answer.\\
Score: 4\\\\
Question: What is the main street in Downtown Naples, Florida's Paradise Coast?\\
Reference Answer: Fifth Avenue South is Downtown's de facto Main Street, a one-mile stretch from 9th Street west to the beach.\\
Candidate Answer: Fifth Avenue South\\
Reasoning: The candidate answer is directly answering the question and is mentioning the same album name as the reference answer. While the reference answer provides additional information about the street those are not needed to answer the question.\\
Score: 4\\\\
\\\#\# Question:\\
<Question>\\
\#\# Reference Answer:\\
<Answer>\\
\#\# Candidate Answer:\\
<Candidate Answer>\\\\
First, explain your assessment of the Candidate Answer based on the criteria above in a few sentences followed by the final score in a new line starting with "\#\# Score:" followed by the score value.}
\end{mdframed}
\caption{Part 2: Prompt used for evaluation of the answer quality.}
\label{fig:answer_eval_prompt_part2}
\end{figure*}

\clearpage
\subsection{Critique-Based Thought Revision Prompt}
\label{sec:appendix_prompts_rag_rethink_cot}
\begin{figure*}[h!]
\begin{mdframed}[backgroundcolor=prompt_background,linewidth=0.0pt]
\texttt{\small
<Previous Thought Generation Steps as Context Instruction-Assistant Turns>\\\\
Your reasoning and answer can be further improved. Below is a critique of your previous reasoning and answer. You should use this critique to improve your reasoning and answer.\\\\
\#\# Critique:
<Critique of Previous CoT>\\\\
Now let's think step by step while considering the provided critique above:\\
Step 1: summarize what is the question is asking, and what are the specific key pieces of information that are needed to answer the question.\\
Step 2: analyze the provided references, one by one. Identify the relevant information that can be used to answer the question. Pay close attention to the entities, names, times, locations, events, and keywords that are relevant to the question. If the question is related to (or implies) the current user location and/or time, you must consider that in finding the relevant information and answering the question.\\
Step 3: based on Step 1 and Step 2, you must provide an answer that directly address the question and is fully grounded on the provided references.\\\\
Now, provide your reasoning steps in a few sentences followed by the final answer to the question as a new line starting with "\#\# Answer:"}
\end{mdframed}
\caption{Prompt used for thought regeneration given previous generated thought and evaluation critique.}
\label{fig:critique_thought_revision_prompt}
\end{figure*}

\subsection{Strategization RAG QA CoT Prompt}
\label{sec:appendix_prompts_strategization}
\begin{figure*}[t]
\begin{mdframed}[backgroundcolor=prompt_background,linewidth=0.0pt]
\texttt{\small
For this task you are asked to answer a question (Question). Please provide factually accurate, direct, and clear responses. Do not ask any clarification questions or ask for additional information.
To make sure the right facts are being considered, you should always ground your responses to the provided references below (References). If the references are not relevant to the question or do not provide the right information, you can respond with an apology rather than fabricating facts.\\\\
\#\# References:\\
<Reference Documents>\\\\
\#\# Question:\\
<Question>\\\\
Before answering the question, take a step back and carefully think about the best strategy to answer the question. Produce an outline for the reasoning steps that you can take to find the best answer. Then, use the outline to think step by step.\\
Use the following template to strategize your reasoning steps, reason step by step, and provide the final answer:\\\\
\#\# Strategy:\\
- Step 1: *** instructions for step 1 ***\\
...\\
- Step N: *** instructions for step N ***\\\\
\#\# Reasoning:\\
- Step 1: *** reasoning corresponding to step 1 in the strategy ***\\
...\\
- Step N: *** reasoning corresponding to step N in the strategy ***\\\\
\#\# Answer: <final answer to the question>}
\end{mdframed}
\caption{Prompt used for CoT strategization to solve the RAG QA task.}
\label{fig:strategization_prompt}
\end{figure*}

\subsection{Distractor Critique Prompt}
The distraction critique process involves evaluating each distractor passage based on three key scores on a scale of 1-5:
    \begin{enumerate}
        \item Relevance Score: Measures how relevant the distractor passage is to the open-ended question, golden passage, location, and time.

        \item Distraction Score: Assesses quality of the passage in its ability to provide relevant retrieval noise.

        \item Format Score: Evaluates the similarity in text length and format between the distractor passage and the original passage.
    \end{enumerate}

The critique process provides constructive feedback on each score, highlighting areas where the distractor passage excels or falls short as a distractor. This feedback is then incorporated into the distractor passage generation step to refine the next iteration of distraction generation.
\label{sec:appendix_prompts_dist_critq}
\begin{figure*}[h!]
\begin{mdframed}[backgroundcolor=prompt_background,linewidth=0.0pt]
\texttt{\tiny
You are an intelligent assistant with expertise in linguistics. Always follow the provided instructions and generate outputs in valid json format without any extra information.\\
\#\#\# User\\
   Given a question, answer, passage, location, user-time, distraction-passage and distraction passage's answer:\\
   1. The goal of the distraction is to provide conflating information from the original passage with respect to the user's question and details, such that it would prompt a human to take a closer look at fine details of both of the passages before coming to an answer. If they glance superficially at a distraction passage, that should feel like a reasonable answer, but when juxtaposed against the original passage, only the original passage should lead to the 'answer'.\\
   2. The goal of distraction-passage is such that when a human is provided distraction-passage and question, they can come to an answer different from the original answer which is retrieved from the passage. Open-question, distraction-answer, distraction-passage should be a slight modification of the original passage question and answer.\\
   3. Open ended question and Distraction-answer is provided along with distraction-passage. Distraction-answer is the answer a human came up with, when provided with just the distraction-passage and the open ended question rather than the original passage and original question. Use that to guide your scoring.\\
    score on the scale of 1 to 5 on
    Scores:\\
       1.relevance-score:\\
            Measures:\\
            a.how relevant the distraction-passage is to the given question, answer, passage, location, user-time. The distraction passage is required to be relevant to the user question, such that when one or two details are omitted from the user question, the distraction passage answers the question sufficiently.\\
       2.distraction:
           Measures:\\
           a. how much of a distraction the passage is to a user who asked the question when provided with both passage and distraction-passage.\\
           b. It ensures that when looking at distraction-passage alone, it would lead to a fully different answer.\\
       3.format:\\
           Measures:\\
           a. how similar in text length and format, the distraction passage is, with respect to the original passage.\\
           b. We require the distraction-passage to be of the same length as the original passage in terms of the number of words, or else the humans can easily distinguish based on the length differences.\\
           c. Ensure format in terms of new lines, spaces etc are similar for original and distraction passage. If there are no new lines, humans can easily distinguish based on format. Penalize omission of new lines if it exists in original but not in distraction.\\
           d. Compute the number of words in original vs distraction-passage and penalize if the difference is noticeable. Penalize if fewer words are present in distraction passage.\\
           d. Penalize format difference, if original has new lines, the distraction should have the same.\\
           e. Ensure if the original has tables, multiple tabs or new lines, the distraction has the same.\\\\
   - Your thought-process field should contain constructive criticism which helps drive improvement. The distraction generation process will display your feedback, so that they can utilize it to provide a better distraction-passage.
   - Your thought process should clearly mention score and reasoning for each type of score and critique for each.\\
   - Think it through step by step and provide a detailed explanation for each score in detail and what all measures are satisfied or missing. Eg: If new lines are missing, it falls under category c,d,e of format.\\
   -  Output in Json with following fields 'relevance-score', 'distraction-score', 'format-score', 'thought-process'.\\
   \#\# question : \{question\}\\
   \#\# answer : \{answer\}\\
   \#\# user-time: \{user-time\}\\
   \#\# location: \{location\}\\
   \#\# passage :\{passage\}\\
   \#\# open ended question: \{open-ended-question\}\\
   \#\# distraction passage: \{distraction\}\\
   \#\# distraction passage's answer: \{distraction-answer\}\\
   \#\#\# Assistant:
}
\end{mdframed}
\caption{Critique prompt used for synthetic distractor generation}
\label{fig:dist_critq_prompt}
\end{figure*}

\subsection{Distractor generation prompt}
\label{sec:appendix_prompts_distractor_generation}
\begin{figure*}[h!]
\begin{mdframed}[backgroundcolor=prompt_background,linewidth=0.0pt]
\texttt{\tiny
You are an intelligent assistant with expertise in linguistics. Always follow the provided instructions and generate outputs in valid json format without any extra information.\\
\#\#\# User :\\
   Think it through step by step:\\
       1. Given a question, answer, passage, location and user-time identify relevant-named-entities, date times, locations in the passage based on the question and answer, such that modifying the relevant named entities will result in a new passage that can cause confusion to the user if they didn't have enough context.\\
       2. You might also be provided with 'prior-distraction-passage', 'prior-distractor-rejecting-reason'. That logs your generation for the same question and answer in the previous turn. Use that information to guide and improve for this round of distraction passage generation.\\
       3.Generate an open ended question "open-ended-question" by modifying provided question such that the answer provided answers the new open ended question.\\
       4. Now using question, answer, user time, location, modify the passage and generate a new passage by modifying in the named entities such that\\
           a. The new passage is relevant to the existing passage.\\
           b.The new passage is grammatically coherent.\\
           c. For the "open-ended-question", both provided and your generated passage are relevant.\\
           d. The distraction-passage should have a similar number of characters as the original passage and similar format.\\
       5. Score your confidence (from 1 to 5) that the generated passage will satisfy condition 4.\\\\
Follow the requirements.\\
\#\# Requirements:\\
   - You must generate the passage based on the user question, location, answer, user-time .\\
   - The generated distraction should be of similar length to the original passage and with similar special characters such as /n, / t. Do not reduce the total number of words.\\
   - Think it through step by step and provide a detailed explanation in the "thought steps" field. Output in Json with following fields 'open-ended-question', 'thought-steps', 'distracting-named-entities', 'distracting-passage', 'score', 'reason' as json fields in your output.\\\\
  Think it through step by step:\\
       1. Given a question, answer, passage, location and user-time identify relevant-named-entities, locations and time information in the passage based on the question and answer, such that modifying the relevant named entities, location or date time  will result in a new passage that can cause confusion to the user if they didn't have enough context.\\
       2. You might also be provided with 'prior-distraction-passage', 'prior-distractor-rejecting-reason'. This provides information on your prior distraction generation for the same question and answer. Use that information along with the 'prior-distractor-rejecting-reason' to analyse and improve the distraction generation and address the gaps.\\
       3.Generate an open ended question "open-ended-question" by modifying the provided question such that the answer provided answers the new open ended question.\\
       4. Now using question, answer, user time, location, modify the passage and generate a new passage by modifying in the named entities and/or location and/or time such that\\
           a. The new passage is relevant to the existing passage.\\
           b.The new passage is grammatically coherent.\\
           c. For the "open-ended-question", both provided and your generated passage are relevant.\\
           d. The distraction-passage should have a similar number of characters as the original passage and similar format.\\
       5. Pick what kind of named entities or location or time change would cause distraction based on user question, opened question , passage and details.\\
           a. Eg: For a query about sporting events today, changing date time of the event will be a distraction as today (datetime) is a key component of the question.\\
           b. Eg: For a question about actors in a movie, changing the movie name slightly and actors name will be a good distraction.\\
       5. Score your confidence (from 1 to 5) that the generated passage will satisfy condition 4.\\\\
}
\end{mdframed}
\caption{Part 1: Prompt used for generating distractor content.}
\label{fig:distractor_gen_prompt}
\end{figure*}

\label{sec:appendix_prompts_distractor_generation_pt_2}
\begin{figure*}[h!]
\begin{mdframed}[backgroundcolor=prompt_background,linewidth=0.0pt]
\texttt{\tiny
\#\# Examples:\\
Example 1 (Named entity change):\\
   question: How hard was the stunt in Mission Impossible 7?
   passage: Mission Impossible 7 was a very dangerous movie. The most dangerous scene was the one where Tom Cruise jumped off a building and held onto a rope. The director was very careful with the stunt and made sure that Tom Cruise was safe."\\
   relevant-named-entities, location, time : Mission Impossible, 7, building
   open-ended-question: What was the most dangerous scene in Mission Impossible?\\
   distracting-named-entities: Mission Impossible, water\\
   distracting-passage: Mission Impossible was a very dangerous movie. The most dangerous scene was the one where Tom Cruise jumped into water and held his breath. The director was very careful with the stunt and made sure that Tom Cruise was safe."\\
   reason: open-ended-question is slight modification of the user question by omitting context that the movie was the 7th Mission Impossible movie. Distracting passage modified both the movie number and the stunt performed. While this would be a correct answer for any mission impossible movie. It's not the correct answer for mission impossible 7, making this the right distractor passage, while being of similar characters and length as the original passage.\\\\
Example 2 (date time change ):\\
   question:  Who are the Washington Commanders playing today?\\
   user-time: 4pm, Thursday, January 9, 2025.\\
   passage:  Title: Washington Commanders vs. San Francisco 49ers - FOX Sports Snippet: Game-time: January 9, 2025, 5pm PST, Venue: Washington D.C, odds: Washington Commanders to win 63
   relevant-named-entities,location, time :  January 9, 2025 5 pm PST, Washington Commanders, San Francisco 49ers\\
   open-ended-question: Who are Washington Commanders playing?\\
   distracting-named-entities, location, time : January 10, 2025, Washington Commanders, New Orleans Saints\\
   distracting-passage: Title: Washington Commanders vs. New Orleans Saints - FOX Sports Snippet: Game-time: January 10, 2025, 5pm PST, Venue: Washington D.C, odds: Washington Commanders to win 52\% probability, Game-a: 0-0 , Game-status: Not-started\\
   reason: open-ended-question made it ambiguous by removing information that the user is asking about today's game (January 9th). The distracting-passage talks about the Washington Commanders vs New Orleans Saints game scheduled on the 10th instead, thus it's a slight modification of the original passage, while answering open ended questions correctly and being a distraction, while being of similar characters and length as the original passage.\\\\
   \#\# question : \{question\}\\
   \#\# answer : \{answer\}\\
   \#\# user-time: \{time\}\\
   \#\# location: \{location\}\\
   \#\# passage :\{passage\}\\
   \#\# prior-distraction-passage: \{prior-distraction-passage\}\\
   \#\# prior-distractor-rejecting-reason: \{prior-distractor-rejecting-reason\}\\
   \#\#\# Assistant:
}
\end{mdframed}
\caption{Part 2: Prompt used for generating distractor content.}
\label{fig:distractor_gen_prompt_2}
\end{figure*}

\end{document}